International Conference on The Enhanced Safety of Vehicles (ESV), Japan, April 2023

SALIENT SIGN DETECTION IN SAFE AUTONOMOUS DRIVING: AI WHICH REASONS OVER FULL VISUAL CONTEXT

**Ross Greer**
**Akshay Gopalkrishnan**
**Nachiket Deo**
**Akshay Rangesh**
**Mohan Trivedi**
Laboratory for Intelligent & Safe Automobiles[1]
University of California San Diego
USA

Paper Number 23-0333

**ABSTRACT**

Detecting road traffic signs and accurately determining how they can affect the driver's future actions is a critical task for safe autonomous driving systems. However, various traffic signs in a driving scene have an unequal impact on the driver's decisions, making detecting the salient traffic signs a more important task. Our research addresses this issue, constructing a traffic sign detection model which emphasizes performance on salient signs, or signs that influence the decisions of a driver. We define a traffic sign salience property and use it to construct the LAVA Salient Signs Dataset, the first traffic sign dataset that includes an annotated salience property. Next, we use a custom salience loss function, Salience-Sensitive Focal Loss, to train a Deformable DETR object detection model in order to emphasize stronger performance on salient signs. Results show that a model trained with Salience-Sensitive Focal Loss outperforms a model trained without, with regards to recall of both salient signs and all signs combined. Further, the performance margin on salient signs compared to all signs is largest for the model trained with Salience-Sensitive Focal Loss.

**INTRODUCTION**

Detecting and recognizing traffic signs is an important module for an autonomous vehicle to observe and interact with its surroundings in a safe manner. The Safety of the Intended Functionality (SOTIF) process [1] examines highly automated systems for possible hazards and triggering events for unintended behaviors; in this framework, failure to detect a sign crucial to driving performance would be considered a triggering event, independent of the hazardous events, based on system limitations. Accordingly, detection systems are continuously improved to push the safe limits of their operation. Until recently, standard object detectors operated by proposing regions of interest or considering a standard set of anchors or window centers within an image, and classifying the contents of the found region. These approaches are typically limited by the span of the convolutional filters which drive them; these filters operate on local windows, or with a pre-determined span and spacing. While the reach of the convolutional filters can be tuned to spread and cover the entire image, doing so creates massive computational costs or creates gaps in coverage. As a solution, the popular transformer model has been proposed as a means of reasoning over the entire image and bringing forward features relevant to the region of interest. To minimize computational costs, this approach has been further refined to include a stage of learning (via a limited number of deformable attention heads) where an image should be sampled to extract meaningful relational features to the region of interest. This approach is known as the Deformable Detection Transformer, introduced in technical detail in the following section.

---

[1] cvrr.ucsd.edu

While advances in detection may improve sign recall, we pose one more consideration to be addressed in driving scenes: many signs simultaneously compete for the attention of a human driver or autonomous driving system. While the ideal intelligent vehicle detection module will have perfect precision and recall of all signs in the field of view, environmental noise and underrepresented examples make it possible that detectors continue to make mistakes. However, on the assumption that error is unavoidable, there are some errors preferred over others. For example, it is less critical that a vehicle passing by a freeway exit sees the sign corresponding to the speed limit of an off-freeway side street, or that a vehicle in the right lane preparing to make a right-hand turn sees the lane guidance for the left lane to navigate the intersection. We ascribe this quality of pertinence and attention-worthiness to the word *salience*, as introduced in [2]. We define the term as follows, with clarification on edge cases further described in Methods section:

**Salience**
A sign is salient if it has the potential to directly influence the next immediate decision to be made by the ego vehicle if no other vehicles were present on the road. Additionally, for signs directing traffic by lane, only signs pertaining to the lane the ego vehicle is in can be classified as salient. In the case of multiple sequential intersections or highway exits visible in the same frame, only signs pertaining to the next immediate intersection or exit could be labeled as salient.

Recent research in sign salience has shown that factors such as sign location, sign appearance, road type, and planned vehicle maneuver can be used to classify signs by salience [2]. Here, we propose a benefit of sign data with salience annotations: salience-aware training methods can be used to improve training of sign detection systems. We make three contributions: (1) creation of the large, salience-annotated LAVA Salient Signs Dataset, (2) definition of Salience-Sensitive Focal Loss, and (3) experimental evaluation of the impact of Salience-Sensitive Focal Loss while training detection transformer models.

**RELATED RESEARCH**

**Traffic Sign Detection and Classification Models**
Traffic sign detection has been well addressed by the field such that near perfect sign detection can be achieved on public sign datasets like the German Traffic Sign Detection Benchmark [3] and similar benchmarks. Detecting traffic signs requires cameras monitoring traffic scenes, which allow us to extract frames from videos and build traffic sign annotation datasets. Trivedi et al. [4] proposed that the best way to capture this traffic surveillance is through a multicamera surveillance approach known as distributed interactive video arrays (DIVA). DIVA helps address issues single view cameras have like handling occlusion and having many overlapping views to obtain 3D information. Such a multicamera system can facilitate easier traffic sign detection by addressing the issues mentioned. Some examples of high performance of traffic sign detection on public traffic sign datasets include:
- Using a separate traffic sign detector model and then a sign recognition model [5]. The traffic sign detection model learns the color of the sign and then the shape, and the sign recognition model works best with an ensemble of CNNs.
- A fully convolutional network to guide traffic sign proposals and then a CNN for sign classification [6]. The FCN learns the rough regions of where the traffic signs are present and the CNN identifies the traffic signs and removes false positives with non-max suppression.
- A Pyramid Transformer that uses atrous convolutions and a RCNN as a backbone [7]. This approach improves the network's ability to detect traffic signs of various sizes.
- Using transfer learning with state-of-the-art object detection models on the German Traffic Sign Detection Benchmark dataset [8]. Faster R-CNN Inception Resnet V2 achieves the best mean average precision while R-FCN Resnet 101 has the best tradeoff between accuracy and execution time.

Transformers have begun to outperform other deep learning techniques like CNNs since they can reason over full image context, or learn where to look to extract more features from an image. The detection transformer DETR [9] is a transformer that allows to learn such global image context and achieves state-of-the-art performance on the COCO object detection dataset. DETR is an end-to-end object detection module that treats object detection as a direct set prediction problem and removes the need for any hand-designed components used by other object detection models. A main weakness of DETR is that it has low performance on detecting small objects. Deformable DETR [10] builds on DETR, reducing the computational complexities and also improving performance on detecting small objects. Deformable DETR uses a different attention module that focuses on a subset of sampling points to perform object detection. This method shows theoretical promise in situations where novel, unusual, or newly emergent signs may appear [11], as the signs can be detected not only on the contents of a box which anchors and tries to recognize the sign's face pattern, but also through inferring on learned generic, face-independent contextual features from training. In this work, we apply this state-of-the-art object detection module to the application of traffic sign detection. In addition, we show that we can steer Deformable DETR to improve performance on salient signs via a novel loss function.

**Traffic Sign Datasets**
There are various traffic sign datasets that allow for researchers to develop traffic sign detection and classifications dataset. A comparison of the size and features of many traffic sign datasets can be seen in [2]. For this paper, we extend the LISA Amazon-MLSL Vehicle Attributes Dataset (LAVA) [12] to create the LAVA Salient Signs (LAVA SS) Dataset, the only dataset which includes the salience property we are interested in utilizing. The datasets and their important properties are listed in the table below:

*Table 1*
***Comparison of Traffic Sign Datasets. The LAVA Salient Signs (LAVA SS) Dataset is used for our research and is the only dataset in this table to include the salience property for traffic signs.***

| Dataset | Number of Images | Important Features |
|---|---|---|
| LISA Traffic Sign Dataset [13] | 7,855 | occlusion, on-side road |
| LISA Amazon-MLSL Vehicle Attributes Dataset [12] | 14,112 | 10s video context, occlusion, salience |
| LAVA Salient Signs Dataset | 31,191 | 10s video context, occlusion, **validated salience** |

**Traffic Object Salience Research**
Learning to focus on salient vehicle objects and construct vehicle visual attention mechanisms has been studied by various researchers, with many using different definitions of what it means to be a "salient" traffic object. We categorize two main types of object saliency from related research: instructive and attentive salience. Attentive salience relates to what objects and directions drivers tend to look at even if these objects may not be what a driver should look at. For this definition of salience, it is often important to monitor the driver's eye gaze to estimate where they are looking at. Taware and Trivedi [14] use driver pose dynamic information to determine the likelihood of a driver gaze zone. This approach tracks facial landmarks like eye corners, nose tip, and nose corners to determine head pose and use the pose to predict the gaze estimation. They found using head pose dynamic features over time increased performance versus using static features like current head pose angles. Robust attentive salience systems must be invariant to different subjects, scales, and perspectives. Vora et al. [15] address this gaze generalization issue using a convolutional neural network to predict driver gaze direction. To improve generalization, they collected a large naturalistic dataset that used ten different subjects and was tested on three unseen subjects. Dua et al. [16] create the first large-scale driver gaze mapping dataset DGAZE, allowing to study attentive salience and where drivers tend to look at for different road and traffic conditions. This dataset contains data from a lab setting of road and driver camera views. Pal et al. [17] learn attentive salience by developing a model named SAGE-Net that uses attention mechanisms to learn how to predict an autonomous vehicle's focus of attention. SAGE-NET uses driver gaze and other important properties like the distance to objects and ego vehicle speed to determine object saliency. Tawari et al. [18] represent gaze behavior for a sequence of image frames by constructing a saliency map using a fully convolutional RNN. The saliency map uses three kinds of pixels: salient (positive) pixels, non-salient (negative) pixels, and neutral pixels. Other than gaze estimation, other important factors like predicting driver maneuvers and braking intent are important to understand attentive salience. Ohn-Bar et al. [19] use a multi-camera head pose estimation model to predict overtaking and braking intent and maneuvers. This system emphasizes real-time performance, which is critical for any attentive salience model in order to timely observe the driver state and react if they are distracted.

In contrast to attentive salience, instructive salience aims to emphasize important objects which an ego vehicle should observe and respond to; these objects should influence the car's future decisions. Our work focuses on instructive saliency and highlights what traffic signs the car needs to be aware of to safely operate. Instructive salience models are often more costly since labeling important objects requires understanding how various road objects and signs influence a driver's decisions and vehicle navigation, so a cognitively-demanding and maneuver-aware manual process is required to annotate such data. To overcome these challenges, Bertasius et al. [20] use an unsupervised learning approach to learn how to detect important objects in first-person images without any instructive salience labels, skipping the costly manual annotation process. The unsupervised network uses a segmentation network to propose possible important objects, and this output is fed into a recognition agent which uses these proposals and other spatial features to predict the important objects. Greer et al. [2] utilize a supervised learning process to classify salient road signs, which can be applied for efficient dataset annotation in future road sign data collection after initial training. Lateef et al. [21] use a conditional GAN to predict what a driver should be looking at in a traffic scene, which parallels this instructive salience definition. For constructing ground truths, they use semantic labels (annotations of traffic objects in images) from various autonomous driving datasets and use various saliency detection algorithms that select which object annotations are the most important. Zhang et al. [22] use interaction graphs to perform object importance estimation in driver scenes. The interaction graph updates features of each object node through interactions with graph convolutions. This task learns to model instructive saliency, as Zhang et al. note that their object importance definition relates to how objects can help with the driver's real-time decision making and improve safe autonomous driving systems.

**METHODS**

**Data Collection**
The LISA Amazon-MLSL Vehicle Attributes (LAVA) dataset contains labeled bounding boxes of traffic signs taken from a front-facing camera of a vehicle. This dataset was collected from the greater San Diego Area and contains a variety of road types, lighting, road types, and traffic conditions. The traffic signs are categorized as stop, yield, do not enter, wrong way, school zone, railroad, red and white regulatory, white regulatory, construction and maintenance, warning, no turn, one way, no turn on red, do not pass, speed limit, guide, service and recreation, and undefined. Along with the traffic sign categorization, we carefully labeled the sign salience property for all the sign annotations. A sign salience validation process was also performed in which the salience property for a sign annotation was checked again for consistency with the above definition of salience; the resulting dataset is referred to as the LAVA Salient Signs (LAVA SS) Dataset. This data collection process ensured that the salience property was properly labeled and the curated dataset had accurate ground truths. In the process of annotation, the provided definition of salience was used as a standard for annotation, with select frequent ambiguous cases handled according to the additional criteria below:
- Guides (signs which indicate the street name, often green) at intersections and freeways were labeled as salient, as long as such signs were the closest such guide in the scene. That is, in the case of multiple sequential intersections, only the guides of the nearest intersection to the car would be labeled as salient. Salient guides should be visible to the vehicle and indicate a possible street the car could turn onto or an exit the car could take. We note that guides tend to have the highest annotator ambiguity, as the class "guide" contains instances of street-level guides as well as freeway-level guides, which may be interpreted differently by different annotators. Likewise, parking guides are a highly missed ground-truth annotation. For this reason, certain applications may benefit from computing precision disregarding guides and parking signs, especially since such signs are less safety critical.
- Instructions pertaining to HOV or Carpool Lanes are marked as salient when the vehicle is moving in the direction of traffic, regardless of lane. Such a sign may indicate a lane available to the intelligent vehicle for optimized traffic flow, or a lane which the vehicle is required to leave if requirements are not met.
- A "No Parking" sign is salient only if the ego vehicle is in a lane which has immediate access to the restricted parking location (e.g. far right lane). An example of this rule is shown in Figure 1.
- While most signs which are facing backwards are marked as non-salient (since they provide instruction to an oncoming lane), in some cases, a yellow reflective warning sign is placed on the back of the sign. In these cases, we mark this as a salient warning sign if the sign is adjacent to the ego vehicle's lane. The vehicle should be aware of such signs to avoid collision with the sign or median. This rule is exemplified in Figure 2.
- Signs which indicate a fine for littering or carpool violations are regarded as non-salient, since an intelligent vehicle should not be littering under any circumstance, nor motivated by the cost of breaking a traffic ordinance.

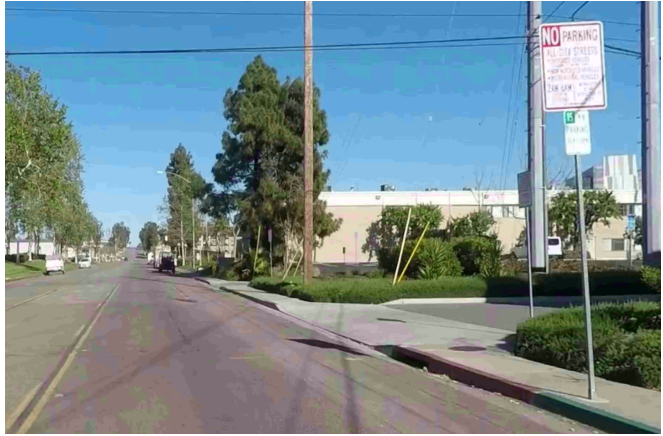

*Figure 1. Because this No Parking sign is located in the lane closest to the ego vehicle, it is considered salient. Were the ego vehicle in the left lane of a two lane road, this would be annotated as non-salient.*

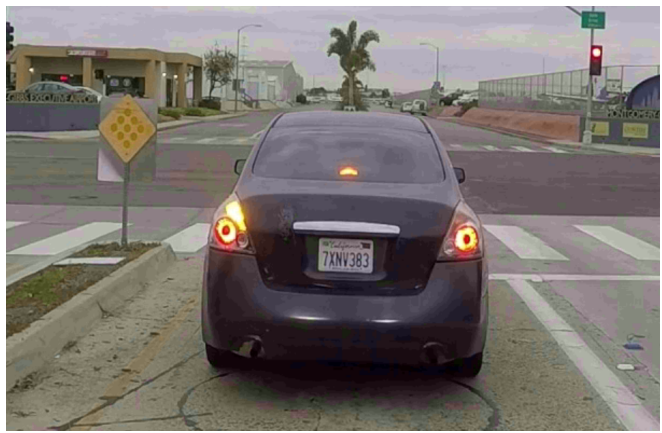

*Figure 2. This sign is facing away from the ego vehicle, but because the reflector is placed to warn the vehicle of its presence, it is annotated as salient.*

Example sign annotations from the LAVA Salient Signs dataset are shown in Figure 3. The LAVA Salient Signs dataset contains 31,992 sign annotations with 20,377 annotations being salient and 11,615 annotations being non-salient. The sign type frequencies for the LAVA Salient Signs Dataset are defined in Figure 4. Because the data was collected and annotated using a selection method which promotes maximal coverage of driving area (including diversity of driving environment, conditions, and road types), the non-uniform distribution of signs may reflects the real-world distribution of salient and non-salient signs as well as the real-world distribution of sign categories during naturalistic driving.

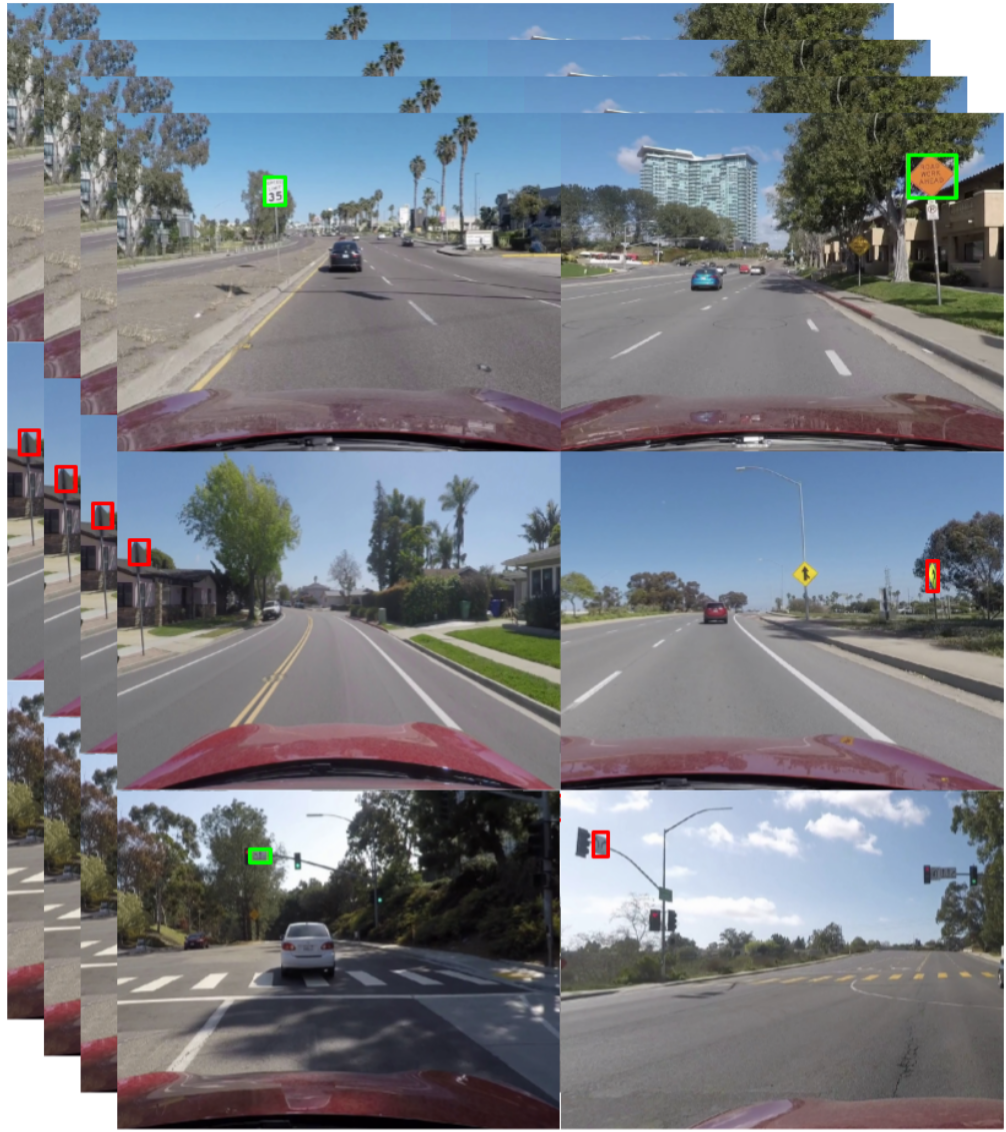

*Figure 3. Example Sign Annotations from the LAVA Salient Signs Dataset. A green bounding box indicates a salient sign and a red bounding box indicates a non-salient sign. As shown in the figure, the salient annotations often mean that the sign relates to the current lane or intersection the driver is in and provides meaningful information that affects the driver's future actions. On the other hand, the non-salient signs are often in a different lane, intersection, or face the wrong way, so these signs don't offer any important information. A vehicle's intended maneuver is important in classifying sign salience, so temporal dynamics should be considered when annotating and utilizing salience data, as explained in [2].*

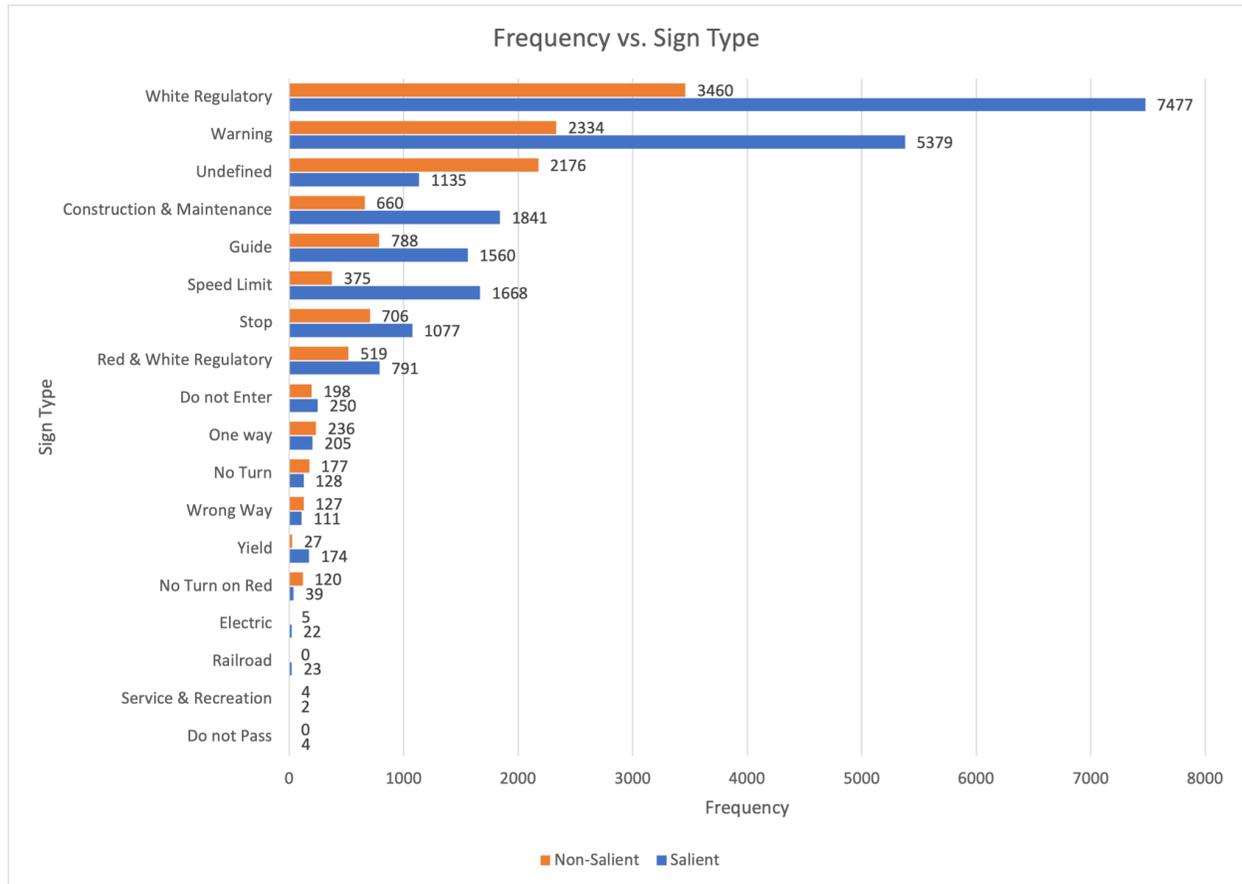

*Figure 4: Sign Type Frequencies in the LAVA Salient Signs Dataset. The blue columns are for salient signs and the orange for non-salient signs. The White Regulatory, Construction & Maintenance, and Warning Signs were the most common sign types. This distribution of signs may be dependent on location, as all of our data was collected in the greater San Diego area.*

**Sign Detection with Deformable DETR**
We use Deformable DETR, introduced in the Related Works section, to detect signs in the images. This detection method forms our performance baseline, described by Figures 2 and 3. We split the LAVA Salient Signs Dataset into 25,591 training instances, 3,200 validation instances, and 3,201 test instances. The model is trained for 15 epochs, retaining the model which reports the strongest precision (with a "hit" at 0.5 intersection-over-union, and 100 maximum detections per image). We use a ResNet50 backbone [23], 300 attention heads, a learning rate of 0.0002, a batch size of 2, and employ gradient clipping and learning rate decay.

**Prioritizing Salient Signs with Salient-Sensitive Loss**
The bounding box regression module of each Deformable DETR detection head is a 3-layer feed-forward neural network. Each detection head also has one linear projection for classification of the estimated bounding box into categories of foreground (object) or background (no object). This classification is trained using a sigmoid focal loss [24], an extension of standard categorical cross-entropy which down-weights easy examples to focus training on hard negatives. The equation for focal loss is

$$FL(p_t) = -\alpha_{FL}(1-p_t)^{\gamma} \log(p_t), \qquad \text{(Equation 1)}$$

where $\alpha_{FL}$ is a hyperparameter to balance the focal loss among other loss functions, $\gamma$ is a focusing parameter to control the influence of hard negatives, and $p_t$ is the predicted probability associated with the ground truth class.

As explained in the introduction, the goal of our detection model is to prioritize successful detection on signs which are salient, ideally placing any model error on non-salient signs. To achieve this, we weigh the focal loss heavily for salient signs according to the function

$$FL(d, p_t) = -\alpha_{FL} w_{SS}(d)(1 - p_t)^\gamma \log(p_t) \qquad \text{(Equation 2)}$$

where $w_{ss}(d) = \alpha_{ss}$ if the ground truth sign nearest detection $d$ is salient, and $w_{ss}(d) = 1$ otherwise. In our case, we use a hyperparameter $\alpha_{ss} = 4$. We name the function $FL(d, p_t)$ *salience-sensitive focal loss*.

**RESULTS**

The performance of the Deformable DETR model on the LAVA Salient Signs Dataset with and without salience-sensitive focal loss is provided in Figures 5-7. These figures display interpolations between precision-recall pairs generated with a detection thresholding of 0 to 1 in increments of 0.1 (with an early stop at thresholds where no positive detections are made). We note that thresholds should be tuned for precision and recall according to intended application; a representative descriptor of performance is given by the precision-recall curves. Results suggest that not only does training with salience-sensitive focal loss distribute error to non-salient signs instead of salient signs, but that the method actually improves overall performance of the model under otherwise equal training.

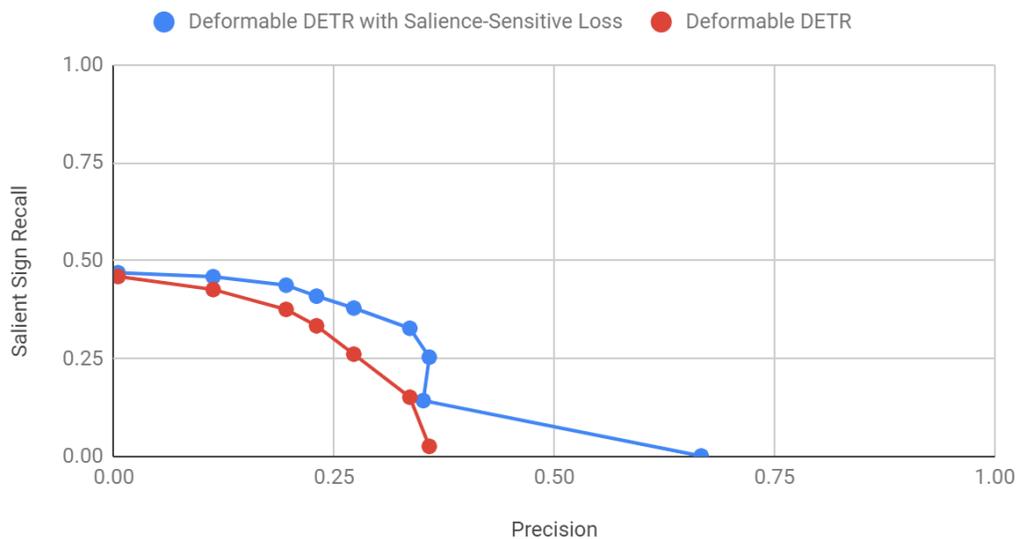

*Figure 5. Deformable DETR shows uniformly better performance in recalling salient signs when using Salience-Sensitive Focal Loss. Additionally, as a general measure of performance, the area under the precision-recall curve is greater when using Salience-Sensitive Loss.*

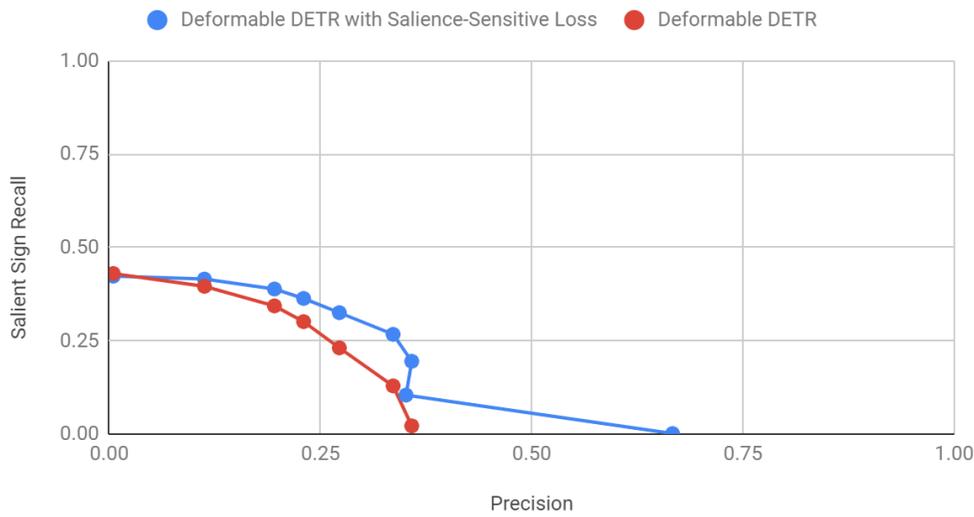

*Figure 6. Deformable DETR additionally shows better performance in recalling all signs (both salient and non-salient) when using Salience-Sensitive Focal Loss. A possible reason for this improvement is that signs which are salient tend to be localized to particular image regions which amass both sign types, whereas some locations of non-salient signs would very rarely have a salient sign appear. This may help guide the transformer as it learns which regions of the image to attend.*

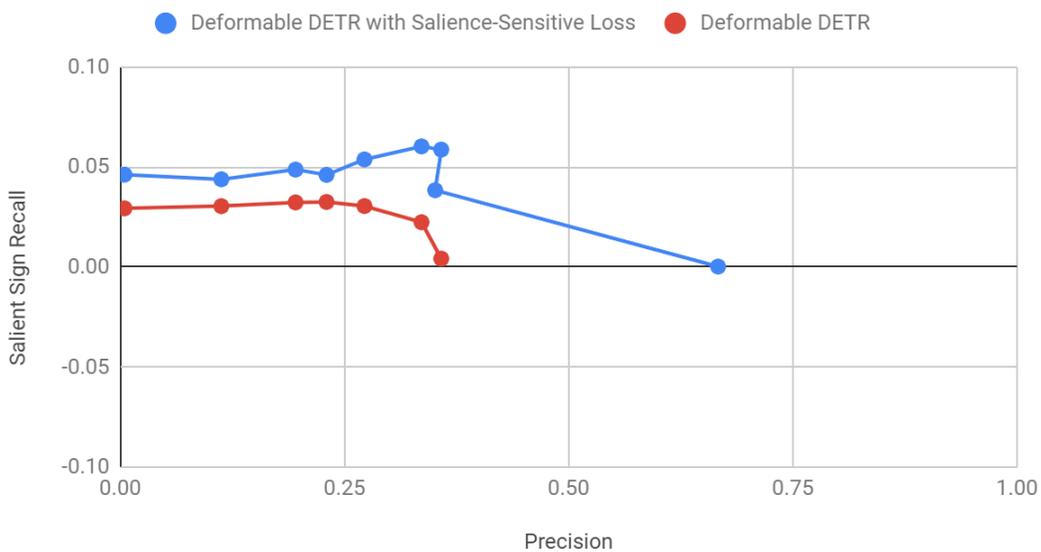

*Figure 7. How well does the Salience-Sensitive Loss bring out performance on salient signs? In this graph, we show the difference in performance between salient sign recall and all sign recall (in other words, how much better is the model at recalling salient signs than the aggregate collection of signs). Deformable DETR does generally perform better on salient signs than all signs together, but with exception as precision increases (in fact, negative at its greatest precision). On the other hand, Deformable DETR with salience-sensitive focal loss maintains improved performance on salient signs, and at greater margin than the baseline model.*

**CONCLUDING REMARKS**

Detection transformers make use of full-image context in a selective manner, and this property makes them an excellent candidate for tasks which often require human drivers to make evaluations over which portion of their visual field to attend to. We illustrated the performance of the recent (and computationally tractable) Deformable DETR model on sign detection for a large dataset even under limited computational budget, providing a baseline for model performance on the dataset. Preliminary results are provided under reduced training time to illustrate the potential of detection-transformer-based methods and to provide a clear demonstration of the impact of modified loss functions on model performance compared to a baseline. Under elongated training regimens and increased dataset sizes, sign detection modules would reasonably be expected to perform to the standards of comparable benchmark models and datasets as described in related research.

We expand this analysis, noting that road objects carry an implicit importance and relevance to the ego vehicle. By including this property, salience, in the training regimen, we show that the sign detector can be further improved, both in general performance and especially in recall of signs which are most important to the safe operation of the autonomous vehicle. Gains in sign detection performance afforded via modification of the training loss function, especially in recalling salient signs, are directly related to the safety of the vehicle in navigating a scene and responding appropriately and safely to surrounding agents.

**ACKNOWLEDGEMENTS**

We are grateful to LISA research sponsors, especially to Dr. Suchitra Sathyanarayana, Ninad Kulkarni, and Jeremy Feltracco of AWS Machine Learning Solutions Laboratory for sharing AWS Sagemaker GroundTruth[2] and LISA colleagues for their assistance in creating the novel LAVA dataset and its derivatives.

---

[2] https://aws.amazon.com/blogs/machine-learning/creating-a-large-scale-video-driving-dataset-with-detailed-attributes-using-amazon-sagemaker-ground-truth/